\pdfoutput=1

\documentclass[11pt]{article}

\usepackage[]{acl}

\usepackage{times}
\usepackage{latexsym}

\usepackage[T1]{fontenc}

\usepackage[utf8]{inputenc}
\usepackage{multirow}
\usepackage{array}
\usepackage{amsmath}
\usepackage{adjustbox}
\usepackage{bm}
\usepackage{graphicx} 
\usepackage{natbib}
\usepackage{caption}
\usepackage{amssymb}
\usepackage{microtype}
\newcommand{\mname}{SixT+ }
\newcommand{\mnamec}{SixT+}
\newcommand{\llr}{$L_{LR}$ }
\newcommand{\llrc}{$L_{LR}$}
\newcommand{\tub}[1]{\underline{\textbf{#1}}}
\newcommand{\sri}{{\rightarrow}}
\newcommand{\sle}{{\leftarrow}}
\newcommand{\slr}{{\leftrightarrow}}
\usepackage{amsmath}

\usepackage{mathtools}
\newcommand{\ronum}[1]{\uppercase\expandafter{\romannumeral #1}}
\usepackage{float}
\usepackage{diagbox}

\usepackage{algorithm}
\usepackage{algorithmicx}
\usepackage{algpseudocode}
\usepackage{booktabs} 
\newcommand{\sx}{\mathbf{x}}
\newcommand{\sy}{\mathbf{y}}

\newcommand{\stheta}{\bm{\theta}}

\newcommand{\mmres}{Ours (m2m)}
\newcommand{\ftall}{XLM-R ft-all}
\newcommand{\mEnres}{Ours (m2En)}
\newcommand{\tabincell}[2]{\begin{tabular}{@{}#1@{}}#2\end{tabular}}

\usepackage{xcolor}

\title{Towards Making the Most of Cross-Lingual Transfer \\ for Zero-Shot Neural Machine Translation }

\author{
  Guanhua Chen$^{1}$\thanks{~~Contribution during internship at Microsoft Research.}~~,
  Shuming Ma$^{2}$,
  Yun Chen$^{3}$\thanks{~~Corresponding author.}~~, \\
 \textbf{Dongdong Zhang$^{2}$, 
  Jia Pan$^{1}$, 
  Wenping Wang$^{4,1}$,  
  Furu Wei$^{2}$ } \\
  $^1$The University of Hong Kong; 
  $^2$Microsoft Research \\
  $^3$Shanghai University of Finance and Economics;
  $^4$Texas A\&M University \\
  \{ghchen,jpan,wenping\}@cs.hku.hk, yunchen@sufe.edu.cn, \\ 
  \{shumma, dozhang, fuwei\}@microsoft.com 
}

\begin{document}
\maketitle
\begin{abstract}
This paper demonstrates that multilingual pretraining and multilingual fine-tuning are both critical for facilitating cross-lingual transfer in zero-shot translation, where the neural machine translation (NMT) model is tested on source languages unseen during supervised training. Following this idea, we present SixT+, a strong many-to-English NMT model that supports 100 source languages but is trained with a parallel dataset in only six source languages. SixT+ initializes the decoder embedding and the full encoder with XLM-R large and then trains the encoder and decoder layers with a simple two-stage training strategy. SixT+ achieves impressive performance on many-to-English translation. It significantly outperforms CRISS and m2m-100, two strong multilingual NMT systems, with an average gain of 7.2 and 5.0 BLEU respectively. Additionally, SixT+ offers a set of model parameters that can be further fine-tuned to other unsupervised tasks. We demonstrate that adding SixT+ initialization outperforms state-of-the-art explicitly designed unsupervised NMT models on Si$\slr$En and Ne$\slr$En by over 1.2 average BLEU. When applied to zero-shot cross-lingual abstractive summarization, it produces an average performance gain of 12.3 ROUGE-L over mBART-ft. We conduct detailed analyses to understand the key ingredients of SixT+, including multilinguality of the auxiliary parallel data, positional disentangled encoder, and the cross-lingual transferability of its encoder.
\end{abstract}

\section{Introduction}
\noindent Neural machine translation (NMT) systems~\cite{sutskever2014sequence,rnnsearch,transformer} have demonstrated superior performance with large amounts of parallel data. However, the performance of most existing NMT systems will degrade when the labeled data is limited~\cite{koehn2017six,Goyal2021TheFE}. To address this problem, unsupervised NMT, in which no parallel corpora are available, is drawing increasing attention. 

Some prior work~\cite{johnson2017google,chen-etal-2017-teacher,Gu2019ImprovedZN,zhang2020improving} use pivot-based methods for zero-shot translation between unseen language pairs. In this setting, both source and target languages have parallel data with a pivot language. However, these approaches are infeasible for rare languages where a parallel dataset of any kind is hard to collect. Another line of work~\cite{Flores,ko2021adapting,GarciaXBT2} build unsupervised NMT through back-translation and further enhance its performance by cross-lingual transfer from auxiliary languages. These methods are usually complicated with multiple iterations of back-translation and a combination of various training objectives. Moreover, their models can only support one or several pre-specified translation directions. Recently, \citet{Chen2021ZeroshotCT} propose SixT, a transferability-enhanced fine-tuning method that better adapts XLM-R \cite{xlmr} for translating unseen source languages. SixT is trained once to support all languages involved in the XLM-R pretraining as the source language. However, they focus on exploring a proper fine-tuning approach and build SixT with the parallel dataset from one auxiliary language, which heavily limits the model's zero-shot translation performance.  

In this paper, we present \mnamec, a strong many-to-English NMT model that can support as many as 100 source languages with parallel datasets from only six language pairs. \mname is trained by applying SixT to multilingual fine-tuning with large-scale data. We first initialize the encoder and embeddings of \mname with XLM-R and then train it with a two-stage training method. At the first stage, we only train the decoder layers, while at the second stage, we disentangle the positional information of the encoder and jointly optimize all parameters except the embeddings. \mname improves over SixT by keeping the decoder embeddings frozen during the whole training process, which speeds up the model training while reducing the model size. \mname is trained once to support all source languages and can be further extended to many-to-many NMT that can support multiple target languages. It is not only a strong multilingual NMT model but can also be fine-tuned for other unsupervised tasks, including unsupervised NMT, zero-shot cross-lingual transfer for natural language understanding (NLU), and natural language generation (NLG) tasks. 

Extensive experiments demonstrate that \mname works remarkably well. For translating to English, \mname significantly outperforms all baselines across 17 languages, including CRISS and m2m-100, two strong unsupervised and supervised multilingual NMT models trained with 1.8B and 7.5B sentence pairs. The many-to-many \mname gets better performance than m2m-100 in 6 out of 7 target languages on the Flores101 testset. When serving as a pretrained model, \mname also performs impressively well. For unsupervised NMT of rare languages, \mname initialization achieves better unsupervised performance than various explicitly designed unsupervised NMT models with an average gain over 1.2 BLEU. For zero-shot cross-lingual transfer for NLU, it significantly outperforms XLM-R on sentence retrieval tasks, while maintaining the performance on most other tasks. On the zero-shot cross-lingual abstractive summarization task, \mname improves mBART-ft by 12.3 average ROUGE-L across 5 zero-shot directions. Finally, we conduct detailed analyses to understand the key ingredients of \mnamec, including multilinguality of the auxiliary parallel data, positional disentangled encoder, and the cross-lingual transferability of its encoder.\footnote{The code and pretrained models are available at \url{https://github.com/ghchen18/acl22-sixtp}.}

\section{\mnamec}
\mname aims at building a strong many-to-English NMT model, especially for the zero-shot directions. We argue that multilingual pretraining and multilingual fine-tuning are both critical for this goal. Therefore, we initialize \mname with XLM-R large and fine-tune \mname on the multilingual parallel dataset with a simple two-stage training method.

\subsection{Data: AUX6 corpus}
We utilize De, Es, Fi, Hi, Ru, and Zh as the auxiliary source languages, which are high-resource languages from different language families. We do not add more auxiliary languages to limit the computation cost and the training data size. The training data is from the WMT and CCAligned dataset, consisting of 120 million sentence pairs. We concatenate the validation sets of auxiliary languages for model selection. We denote this dataset as \emph{AUX6}. More dataset details are in the appendix. Following \citet{conneau2019cross}, sentences of the $i^{\text{th}}$ language pair are sampled according to the multinomial distribution calculated as follows:  
\begin{equation}
    q_{i}= \frac{p_{i}^{\alpha}}{\sum_{j} p_{j}^{\alpha}},
\end{equation}
where $p_j$ is the percentage of each language in the training dataset and we set the hyper-parameter $\alpha$ to be $0.2$. In all experiments, all texts are tokenized with the same sentencepiece \citep{sentencepiece} tokenizer as XLM-R.

\begin{figure*}[t]
  \center
  \includegraphics[width=0.95\textwidth]{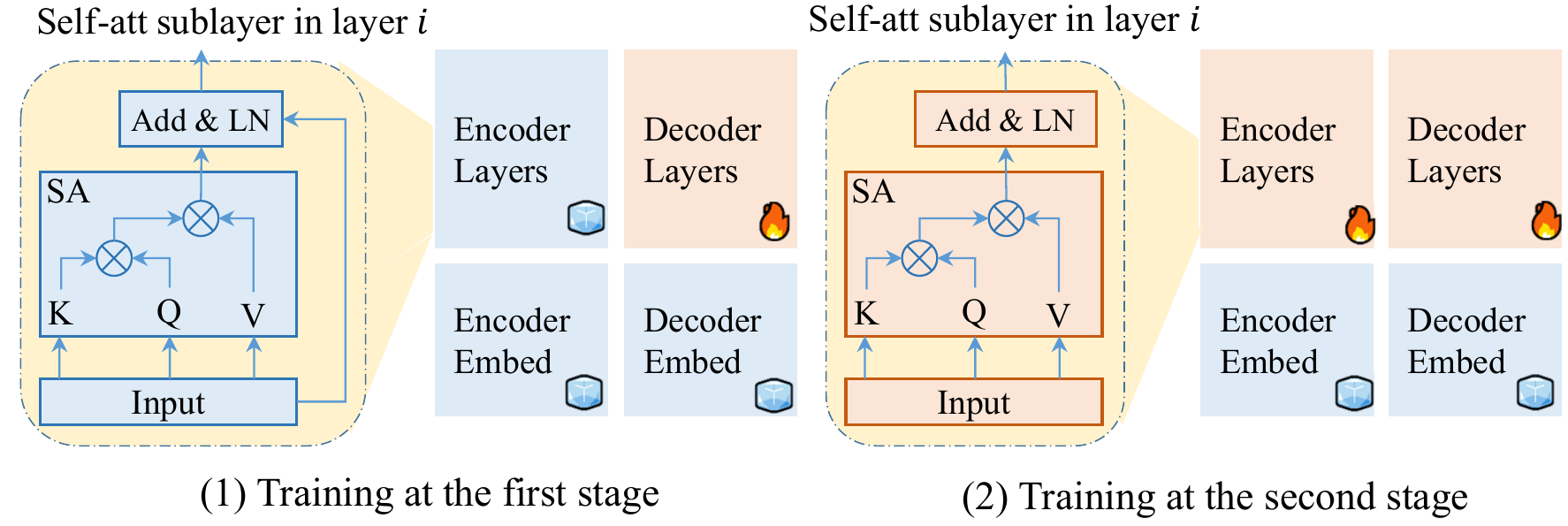}
  \caption{Our proposed two-stage training framework (TransF) for building cross-lingual NLG model with XLM-R. The blue icy blocks are initialized with XLM-R and frozen, while the red fiery blocks are initialized randomly or from the first stage. `SA' denotes the self-attention sublayer. We remove the residual connection at the 23$^\text{th}$ (penultimate) encoder layer at the second stage, namely $i=23$ in the figure.}
  \label{fig:train_method}
\end{figure*}

\subsection{Model}

\paragraph{Architecture} \mname is a Transformer-based NMT model with $\sim$0.7B model parameters. To initialize the encoder with XLM-R large, our encoder has the same configuration as XLM-R large, i.e., 24 encoder layers, hidden state dimension of 1024, feed-forward dimension of 4096, and head number of 16. For the decoder, we follow the suggestion in \citet{Chen2021ZeroshotCT}, which has 12 decoder layers, a hidden state dimension of 1024, feed-forward dimension of 3072, and head number of 16. We use the same vocabulary as XLM-R and tie the encoder embeddings, decoder embeddings, and decoder output projection to reduce the model size.

\paragraph{Learning} We first initialize the encoder and embeddings with XLM-R large and then fine-tune the model on the auxiliary parallel dataset. Compared with fine-tuning XLM-R for NLU tasks like text classification, the prediction space for \mname is much larger and it has to learn much more randomly initialized parameters. Directly fine-tuning all parameters may degrade the cross-lingual transferability which is learned in XLM-R. Therefore, following \citet{Chen2021ZeroshotCT}, we train \mname with a two-stage training framework, as shown in Figure~\ref{fig:train_method}.
\paragraph{Stage 1: Decoder Training.} To preserve the cross-lingual transferability of XLM-R, we first train the decoder by keeping the encoder frozen:  
\begin{align}
        \mathcal{L}_{\stheta_{\text{dec}}}=\sum_{D_i \in D}\sum_{\langle \sx,\sy \rangle \in D_i} \log P(\sy|\sx;  ,\stheta_{\text{dec}}),
\end{align}
where $D = \{D_1;...;D_K\}$ is a collection of parallel dataset in $K$ auxiliary languages, $\langle \sx,\sy \rangle$ is a parallel sentence pair with source language $i$ and $\stheta_{\text{dec}}$ is the parameter set of the decoder layers. 
\paragraph{Stage 2: Fine-tuning.} Freezing the encoder parameters limits the NMT model capacity, especially for the large-scale training data. Therefore, we jointly train the full model in another stage:
\begin{align}
   \mathcal{L}_{\stheta}=\sum_{D_i \in D}\sum_{\langle \sx,\sy \rangle \in D_i} \log P(\sy|\sx;\stheta),
\end{align}
where $\stheta$ is the parameter set of both encoder and decoder layers. 

Different from SixT which fine-tunes the decoder embedding, we keep the embeddings fixed during the whole training process (see Figure~\ref{fig:train_method}). Our preliminary experiments find that this strategy leads to higher computational efficiency without degrading the performance.

\paragraph{Positional Disentangled Encoder} Positional Disentangled Encoder (PDE) is reported to improve zero-shot NMT in the previous work \cite{liu2020improving,Chen2021ZeroshotCT}. The positional correspondence between the input tokens and the encoder representations is one of the factors that makes the encoder representations language-specific. PDE relaxes such correspondence by removing residual connections in an encoder layer. We refer the readers to \citet{liu2020improving,Chen2021ZeroshotCT} for more details. In \mnamec, we remove the residual connection after the self-attention sublayer of the 23$^\text{th}$ (penultimate) encoder layer at the second training stage, as suggested by \citet{Chen2021ZeroshotCT}. For simplicity, we denote the two-stage training method with PDE as \emph{TransF} in the following sections.

\section{Zero-Shot Neural Machine Translation}
\begin{table*}[t]
  \centering 
  \resizebox{1.0\textwidth}{!}{
  \begin{tabular}{l|ll|rr|rrr|rrr|rr|rr}
    \toprule 
    \multirow{2}{*}{Model}  & \multirow{2}{*}{\# Sent} & \multirow{2}{*}{Param.} & \multicolumn{2}{c|}{German} & \multicolumn{3}{c|}{Romance} & \multicolumn{3}{c|}{Uralic} & \multicolumn{2}{c|}{Slavic} & \multicolumn{2}{c}{Arabic}  \\ 
                            &      &  &  De   & Nl    & Es    & Ro    & It    & Fi    & Lv    & Et    & Ru   & Pl    & Ar    & Ps    \\ \midrule
    CRISS                & 1.8B  & 0.6B & 28.8  & 47.0  & 32.2  & 35.4  & 48.9  & 23.9  & 18.6  & 23.5  & 21.2 &  $-$  & 28.2  & $-$  \\ 
    m2m-100             & 7.5B & 1.2B & 31.9  & 54.0  & 32.8  & 38.3  & 55.9  & 29.0  & 23.0  & \tub{30.7}   & \tub{24.2} & \tub{29.9}  & 28.4  & 10.9    \\ 
    SixT                & 0.04B & 0.7B & 33.8 & 54.7 & 30.1 & 33.9 & 43.0 & 26.3 & 19.7 & 25.7  & 20.4 & 23.9 &25.1 &11.4   \\  
    mBART-ft             & 0.12B & 0.6B & 32.2  & 50.6  & 33.0  & 34.0  & 53.3  & 28.7  & 17.9  & 22.0  & 21.7 &  15.0 &  19.2 &  0.9    \\  
    XLM-R ft-all            & 0.12B  & 0.7B & 32.8  & 37.7  & 34.4  & 32.5  & 37.2  & 29.5  & 17.9  & 23.7  & 23.4 &  19.6 &  22.3 &  8.5    \\ \midrule 
    \mname (1st)  & 0.12B & 0.7B & 33.7  &  52.5  &  34.1  &  36.8  &  49.4  &  30.0  &  21.4  &  27.4  &  22.3  &  25.7  &  27.3  &  12.2    \\
    \mname              & 0.12B & 0.7B & \tub{35.3}  &  \tub{58.5}  &  \tub{35.2}  &  \tub{38.6}  & \tub{60.9}  &  \tub{32.1}  &  \tub{23.3}  &  30.5  &  \tub{24.2}  &  28.1  &  \tub{30.5}  &  \tub{14.9} \\ 
  \end{tabular}}
  \resizebox{1.0\textwidth}{!}{
  \begin{tabular}{l|rrrr|rr|rrr|rr|rr}
    \toprule 
    \multirow{2}{*}{Model}  &  \multicolumn{4}{c|}{Indo-Aryan} & \multicolumn{2}{c|}{Turkic} & \multicolumn{3}{c|}{East Asian} & \multicolumn{2}{c|}{Khmer} &  \multicolumn{2}{c}{\multirow{2}{*}{Avg.}} \\ 
                           &          Hi   & Ne    & Si    & Gu    & Tr    & Kk    & Zh    & Ja    & Ko    & My    & Km    &       \\ \midrule
      CRISS                  & 23.1 & 14.7  & 14.4  & 19.0  & 20.6  &  10.1  & 13.4  &  7.9  & 24.8  &  6.7  & $-$   & $-$ &23.1$^{\dagger}$  \\ 
      m2m-100                & 24.5 &  5.2  & 15.3  &  0.5  & 25.5  &  2.1  & \tub{23.8}  & \tub{13.9}  & \tub{36.1}  &  2.0  &  6.7  & 23.7 & 24.9$^{\dagger}$   \\ 
      SixT & 17.5 & 14.4 & 12.2 & 17.3  & 21.7 &19.0 & 13.4 & 10.7 & 31.2 & 5.4 &9.8 &22.6 & 23.8$^{\dagger}$ \\ 
      mBART-ft                 & 25.7 &  18.0 &  8.8  & 15.4  & 21.2  & 19.6  & 19.3  & 10.0  & 30.7 &  3.6 &  0.1 &  21.8 & 24.2$^{\dagger}$  \\ 
      XLM-R ft-all               & 27.6 &  19.9 &  10.4 &  18.2 &  20.1 &  20.7 &  19.3 &  9.5  & 16.6 &  4.1 &  8.4 &  21.5 & 22.9$^{\dagger}$  \\ \midrule
      \mname (1st)             &27.3  &  20.4  &  14.7  &  23.9  &  23.3  &  23.3  &  19.3  &  10.8  &  24.8  &  10.4  &  10.3 & 25.3 & 26.7$^{\dagger}$       \\ 
      \mname              & \tub{29.8}  &  \tub{23.7}  &  \tub{17.5}  &  \tub{27.5}  &  \tub{27.5}  &  \tub{27.3}  &  21.6  &  13.1  &  33.3  &  \tub{15.3}  &  \tub{12.5}      & \tub{28.7}  &\tub{30.3}$^{\dagger}$    \\ 
      \bottomrule
      \end{tabular}}
    \caption{BLEU comparison with baselines on many-to-English test sets. `\# Sent' is the training data size. `Param.' is the model size. `$-$' indicates the language is not supported by CRISS. ${\dagger}$ is the average BLEU across the source languages supported by CRISS. \mname (1st) is the \mname after the first training stage. The best BLEU is bold and underlined. The last three utilize the same multilingual pretrained language model (XLM-R large) but with a different fine-tuning method.} 
    \label{tab:main-res}
\end{table*}
\subsection{Experiment Settings} 
For the many-to-English translation task, we evaluate the performance of \mname on the test sets of 23 language pairs from 9 various language groups\footnote{We refer to the language groups information in Table 1 of \citet{fan2020m2m}.}: German group (De, Nl), Romance group (Es, Ro, It), Uralic and Baltic group (Fi, Lv, Et), Slavic group (Ru, Pl), Arabic group (Ar, Ps), Indo-Aryan group (Hi, Ne, Si, Gu), Turkic group (Tr, Kk), East Asian group (Zh, Ja, Ko) and Khmer group (My, Km). The dataset details are in the appendix. For decoding, we use beam-search with beam size 5 for all translation directions and do not tune length penalty. We report detokenized BLEU for all directions using sacrebleu\footnote{BLEU+case.mixed+numrefs.1+smooth.exp+tok.13a\\+version.1.5.0}.

We compare \mname with SixT and four other baselines. Among the four baselines, XLM-R ft-all and mBART-ft use the same training data as \mnamec, while CRISS and m2m-100 are trained on 1.8B and 7.5B sentence pairs. As \mnamec, CRISS, and m2m-100 have different model sizes, support different numbers of languages and are trained with different training datasets, the comparisons are not completely fair, but the results can still demonstrate the strong performance of \mnamec. 

\noindent$\bullet$ CRISS \citep{tran2020criss}. This model is the state-of-the-art unsupervised many-to-many multilingual NMT model. It is initialized with mBART and fine-tuned on 180 translation directions from CCMatrix. It only supports 25 input languages. 

\noindent$\bullet$ m2m-100 \citep{fan2020m2m}. This model is a strong supervised many-to-many multilingual NMT model. It is a large Transformer trained on huge parallel data across 2200 translation directions and with 7.5B parallel sentences from CCMatrix and CCAligned as well as additional back-translations. The official 1.2B model is evaluated.

\noindent$\bullet$ SixT \citep{Chen2021ZeroshotCT}. This model motivates \mnamec. The SixT model trained with XLM-R large on WMT19 De-En is evaluated and compared. 

\noindent$\bullet$ mBART-ft \citep{mbart,tang2020multilingual}. mBART\footnote{We use mBART50 from \citet{tang2020multilingual}.} is a strong pretrained multilingual seq2seq model. We follow their setting and directly fine-tune all model parameters on the AUX6 corpus. 

\noindent$\bullet$ XLM-R ft-all \citep{conneau2019cross}. This method is the same as \mname but utilizes a different fine-tuning method that directly optimizes all model parameters.
\begin{table*}[!t]
  \centering 
  \resizebox{0.62\textwidth}{!}{
    \begin{tabular}{l|rrrrrrrr}
    \toprule
    Target Lang.  & En    &  De    &  Es          &  Fi          &  Hi    & Ru  &  Zh  & Avg. \\ \midrule
    m2m-100       & 23.6  &  15.9  &  15.2  &  \tub{11.3}  &  14.1  &  14.3  &  19.9  &  16.3 \\
    \mname m2m  & \tub{29.8} & \tub{17.4}  & \tub{15.3}  & 10.2 &  \tub{15.5}  &  \tub{14.6}  &  \tub{25.2}  &  \tub{18.3}  \\
    \bottomrule
    \end{tabular}}
  \caption{Averaged BLEU comparison of \mname m2m and m2m-100 on zero-shot translations. The detailed results are in the Table~\ref{tab:app-m2m-task-results} of the appendix.} 
  \label{tab:task-results}
\end{table*}

\subsection{Main Results}  
As shown in Table~\ref{tab:main-res}, \mname outperforms all baselines with an average gain of 5.0-7.2 BLEU. The performance of \mname is impressive given that it does not use any other monolingual or parallel texts except the 0.12B parallel sentence pairs. First, the significant improvement over mBART-ft demonstrates that the multilingual pretrained encoder XLM-R can also build a strong zero-shot many-to-one translation model if fine-tuned properly. Second, \mname is significantly better than XLM-R ft-all and \mname (1st), proving that a proper fine-tuning method is important for zero-shot translation. Finally, the gain of \mname over SixT shows that adding more auxiliary languages and more parallel data benefits the performance. 

\mname achieves new state-of-the-art performance on unsupervised many-to-English translation. It is significantly better than CRISS in all $14$ unsupervised directions. When comparing with supervised models, \mname improves over m2m-100 on $17$ out of $23$ translation directions. Although CRISS and m2m-100 are many-to-many NMT models that may face the \emph{insufficient modeling capacity} problem \citep{zhang2020improving}, they are strong many-to-English baselines trained with much more data (1.8 billion for CRISS and 7.5 billion for m2m-100) and computation cost. Moreover, the model size of m2m-100 is much larger than \mnamec. 

Different from previous unsupervised NMT models built with back-translation on monolingual data~\citep{lample2018unsupervised,lample2018phrase} or parallel data mining~\citep{tran2020criss}, \mname illustrates that better unsupervised NMT can be achieved by cross-lingual transfer from auxiliary languages. It improves on the test sets whose languages are in the same family as the auxiliary languages. For languages that are not in the same family of auxiliary languages, \mname also works well. For instance, it improves My$\sri$En from 6.7 to 15.3 BLEU, Ps$\sri$En from 10.9 to 14.9 BLEU, and Kk$\sri$En from 20.7 to 27.3 BLEU. 

\subsection{Analysis}\label{sec:ana}
\paragraph{Many-to-Many \mname}
The \mname can be extended to support other or multiple target languages. Following \citet{zhang2020improving}, we build a many-to-many \mname (\mname m2m) model and switch between different target languages by a target-language-aware linear projection layer between the encoder and the decoder. The linear layers are randomly initialized and trained in both training stages. The model is also trained on AUX6, but additionally includes the En$\sri$\{De,Es,Fi,Hi,Ru,Zh\} translation directions during supervised training and validation. All the other training details are the same. We evaluate the performance of \mname m2m on the Flores 101 testset~\cite{Goyal2021TheFE}, which is a multilingual aligned benchmark that covers 101 different languages. Following previous work~\cite{fan2020m2m}, we report tokenized BLEU when Hindi\footnote{\url{https://github.com/anoopkunchukuttan/indic_nlp_library}} and Chinese\footnote{We use the default Chinese tokenizer of sacrebleu.} are the target language and the detokenized BLEU for other target languages. We compare it with the m2m-100 (1.2B) model, as shown in Table~\ref{tab:task-results}. Detailed results on each source language are in Table~\ref{tab:app-m2m-task-results} of the appendix.

Overall, our model outperforms m2m-100 in 6 out of 7 target languages. This is impressive given that our model is unsupervised. The \mname m2m performs more evenly in different source languages (see Table~\ref{tab:app-m2m-task-results} in the appendix). In contrast, the performance of m2m-100 varies across languages. Our model learns to translate through effective cross-lingual transfer, while m2m-100 relies heavily on the scale and quality of the direct parallel dataset. We also compare \mname m2En and \mname m2m for translating to English on this testset and get an average BLEU of 30.5 and 29.8, respectively (see Table~\ref{tab:app-m2m-task-results} in the appendix). The results demonstrate that \mname m2m successfully supports seven target languages while keeping most of the performance of \mname m2En on the many-to-English testset.
\begin{table}[t]
  \centering 
  \resizebox{0.95\columnwidth}{!}{
  \begin{tabular}{l|l|rrrrrr}
  \toprule
   Data    &  Size  &  Hi  & Ne & Si & Gu & Avg.  \\ \midrule
   De-En   &  8M & 17.3  &  13.7  &  11.9  &  16.0  & 14.7   \\
  4 Aux. Langs &  8M & 20.9  &  16.6  &  15.1  &  20.9  & 18.4   \\
  \bottomrule
  \end{tabular}}
   \caption{BLEU comparison of \mname trained with the same size of training data that consists of different number of auxiliary languages. `4 Aux. Langs' is a combination of \{De,Es,Fi,Ru\}-En parallel datasets. }
  \label{tab:dif-lang-diverse-result}
\end{table}
\begin{table}[t]
  \centering 
  \resizebox{0.9\columnwidth}{!}{
  \begin{tabular}{l|ccc}
  \toprule  
  \multirow{2}{*}{Data}  &   Europarl   &  WMT19  & AUX6  \\
    &   (1.9M)   &  (41M)  &  (120M)  \\ \midrule
   \mnamec{}     &   21.5  &  26.3  &  32.9  \\
   \mnamec{} w/o PDE  &   20.5  &  26.1  &  32.9  \\
  \bottomrule
  \end{tabular}}
  \caption{The average BLEU of \mname with and without positional disentangled encoder (PDE). Note that AUX6 includes more source languages. The detailed scores are in the Table~\ref{tab:app-pos-enc} of the appendix. }
  \label{tab:pos-enc}
\end{table}
\begin{table*}[t!]
\centering
  \resizebox{0.8\textwidth}{!}{
  \begin{tabular}{l rrrrrrrr}
  \toprule
  & \textbf{XNLI} & \textbf{PAWS-X} & \textbf{POS} & \textbf{NER}  & \textbf{MLQA}  & \textbf{BUCC} & \textbf{Tatoeba} & \textbf{Avg.} \\
   \textbf{Metric} & \textit{acc.} & \textit{acc.} & \textit{F1} & \textit{F1} & \textit{F1 / EM} & \textit{F1} & \textit{acc.} & -- \\
   \textbf{\# langs.} & 15 & 7 & 33 & 40  & 7  & 5 & 37 & -- \\  \midrule
    Vanilla XLM-R          & 79.2  &  86.4  &  74.2  &  65.4  &    71.6 / 53.2   &  66.0  &  57.7  &  71.5  \\
    XLM-R FT-all           & 75.9  &  85.9  &  67.1  &  52.1  &    62.9 / 44.0   &   7.9  &  59.5  &  58.8      \\
    Ours (m2En)            & 78.5  &  88.0  &  76.1  &  62.2  &    68.7 / 48.9   &  85.9  &  81.4  &  77.3      \\ 
    Ours (m2m)             & 80.0  &  88.3  &  74.4  &  59.0  &    70.7 / 51.7   &  88.0  &  81.4  &  77.4   \\ \midrule
   \citet{phang2020english} & 80.0 & 87.9 & 74.4 & 64.0 &  72.4 / 53.7 & 71.9 & 81.2 & 76.0 \\
  \bottomrule
  \end{tabular}}
  \caption{XTREME benchmark results of our models and baselines. The results for individual languages can be found from Table~\ref{tab:full_xnli} to Table~\ref{tab:full_tatoeba} in the appendix.
  }
\label{tab:xtreme_results}
\end{table*}
\paragraph{Effect of the Multilinguality of Auxiliary Languages}
Previous studies report that adding more parallel data and more auxiliary languages improves performance for unsupervised NMT~\cite{Garca2020AMV,Bai2020UnsupervisedNM,GarciaXBT2}. In this experiment, we examine whether increasing multilinguality under a fixed data budget improves the zero-shot performance of \mnamec. We fix the amount of auxiliary parallel sentence pairs to 8 million and vary the number of auxiliary languages. We report the results in Table~\ref{tab:dif-lang-diverse-result}. It is observed that the model trained with four auxiliary languages (De, Es, Fi, Ru, each has the same data size) outperforms that of one auxiliary language (De), with an average gain of 3.7 BLEU. Note that for both cases, we use auxiliary languages which are not in the Indo-Aryan group to remove the impact of language similarity. This observation demonstrates the necessity of utilizing multiple auxiliary languages in the training dataset. 

\paragraph{Effect of Positional Disentangled Encoder}
In this part, we conduct a comprehensive study on the effect of the positional disentangled encoder (PDE) \cite{liu2020improving,Chen2021ZeroshotCT}. Table~\ref{tab:pos-enc} presents the results. We find that on the small-scale Europarl dataset, PDE improves the zero-shot performance with an average gain of 1.0 BLEU. However, when the training data goes large or/and becomes more multilingual, the gain decreases (see results on WMT19 and AUX6). To confirm this, we also conduct experiments on \mname m2m (see Table~\ref{tab:app-m2m-task-results} in the appendix). For translating to English, the models with and without PDE perform comparably well. However, for translating to other languages, PDE improves in 5 out of 6 directions, with an average gain of 0.4 BLEU. This is expected as these directions include only one source language (En) and much less training data (7M$\sim$41M) than translating to English (120M). In summary, when large-scale multilingual training data are available for all target languages, it is fine to remove PDE. We suspect the model has already learned language-agnostic encoder representations in this case. Otherwise, PDE benefits zero-shot performance. 

\paragraph{Performance on Cross-lingual NLU Tasks}
To better understand the encoder representation produced by \mnamec, we evaluate the zero-shot cross-lingual transfer performance of the \mname encoder on the XTREME benchmark \cite{hu20xtreme}. The XTREME includes 9 target tasks of natural language understanding. We do not report results on XQuAD and MLQA as they have no held-out test data~\cite{phang2020english}. For all other XTREME tasks, we follow the training and evaluation protocol in \citet{hu20xtreme} and implement with the jiant toolkit~\cite{phang2020english}. As NMT training can be regarded as an intermediate task~\cite{pruksachatkun2020intermediate}, we include previous results on using English intermediate NLU tasks to improve XLM-R on XTREME as a reference~\cite{phang2020english}. Table~\ref{tab:xtreme_results} provides the average results for each task. The detailed results are in the appendix. 

Overall, \mname encoders achieve 8.3\% and 31.6\% performance gain over XLM-R and XLM-R ft-all across the seven tasks, which verifies that our model learns a more language-agnostic encoder representations. Our encoder may learn better sentence-level representation and capture better semantic alignments among parallel sentences through multilingual NMT training, therefore it generally performs better on sentence pair (XNLI and PAWS-X) and sentence retrieval tasks (BUCC and Tatoeba). The results show the potential of leveraging NLG task as the intermediate task for improving performance on XTREME. We leave a more detailed exploration of why NMT training as well as other NLG intermediate tasks could be beneficial for a given NLU task as future work.

\section{\mname as a Pretrained Model}
\mname learns language-agnostic encoder representation and performs impressively well on translating various source languages. In this part, we extend \mname to two cross-lingual NLG tasks where the direct labeled data is scarce, namely unsupervised NMT for low-resource languages and zero-shot cross-lingual abstractive summarization.

\subsection{Unsupervised NMT for Low-resource Language}
Given a low-resource language pair where the parallel dataset is unavailable, early work on unsupervised NMT build the translation model by training denoising autoencoding and back-translation concurrently~\cite{lample2018phrase,lample2018unsupervised,artetxe2018unsupervised}. However, these methods may lack robustness when languages are distant~\cite{kim2020and,marchisio2020does}. For example, \citet{guzman2019flores} report BLEU scores of less than 1.0 on distant language pair Nepali-English using the method in \citet{lample2018phrase}. Recent work improves by better initializing the unsupervised NMT model either with a multilingual pretrained language model~\cite[MulPLM]{mbart,song2019mass,ko2021adapting} or a multilingual NMT model~\cite{lin2020pre}. In this part, we follow this line and offer an alternative initialization option for building strong unsupervised NMT models. 

We first initialize the \llrc$\sri$En model with \mnamec. As \mname only supports En as the target language, we initialize the En$\sri$\llr model with XLM-R following how \mname is initialized. Then we iteratively improve these two models with back-translation. For simplicity, we do not update the \llrc$\sri$En model and only train the reverse model once. We train it with a synthetic back-translation dataset from \llr monolingual data using the two-stage training method\footnote{We do not use PDE here as PDE may harm the supervised performance of the reverse model.}. We do not apply other unsupervised NMT techniques, such as iterative back-translation~\cite{lample2018phrase}, cross-translation~\cite{GarciaXBT2} or iterative mining of sentence pairs~\cite{tran2020criss}. These methods could be complementary to our method. We leave the in-depth exploration as future work. 

\paragraph{Experimental Settings}
\begin{table}[t]
  \centering 
  \resizebox{1.0\columnwidth}{!}{
  \begin{tabular}{llrrrrr}
  \toprule
    \multirow{2}{*}{ID} & \multirow{2}{*}{Method}  &  \multicolumn{2}{c}{Ne-En}    & \multicolumn{2}{c}{Si-En}    \\ 
                          &    & $\sri$  &   $\sle$   &  $\sri$    &   $\sle$   \\  \midrule
  \multicolumn{2}{l}{\em{Supervised approach}}  &   &  &  \\
 (1) & m2m-100              &   5.2   &  0.4  &  15.3  & 4.6   \\ 
 (2) & \citet{Flores}$^\dagger$          &  \underline{21.5}   &  8.8  &  15.1  & 6.5   \\
 (3) & \citet{mbart}$^\dagger$             &  21.3   &  \underline{9.6}  &  \underline{20.2}  & \underline{9.3}   \\\midrule 
  \multicolumn{2}{l}{\em{Unsupervised approach}}  &  &  &  &  \\
  (4) & CRISS & 14.7   &  5.5  &  14.4  & 6.0   \\ 
  (5) &\citet{Flores} $^\dagger$         & 18.8   &  8.3  &  $-$  & $-$   \\
 (6) & \citet{ko2021adapting} $^\dagger$         & 18.8   &  9.2  &  $-$  & $-$  \\
  (7) & \citet{GarciaXBT2}  $^\dagger$         & 21.7   &  8.9  &  16.2  & 7.9   \\ 
  (8) & Ours (\mnamec)             & \textbf{23.7}   & \textbf{10.1}  &  \textbf{17.5}  & \textbf{8.2}   \\
  \bottomrule
  \end{tabular}}
  \caption{BLEU comparison of different models on the low-resource language translation. Results with `$\dagger$' are quoted from the original paper. The best unsupervised method for each translation direction is bold, while the best supervised method is underlined.}
  \label{tab:unsup-result}
\end{table} 
We evaluate our method on Ne and Si, two commonly used benchmark languages for evaluating low-resource language translation. The monolingual dataset of Ne and Si consists of 7 million sentences that are sampled from CC100 and CCNet dataset. The test sets are from the Flores dataset \cite{Flores}. We set the beam size to 5 during the offline back-translation and select the model with unsupervised criterion in ~\citet{lample2018unsupervised}. We compared with state-of-the-art supervised and unsupervised baselines. Please refer to the appendix for more details.

\paragraph{Results}  The results are illustrated in Table~\ref{tab:unsup-result}. Our model outperforms all unsupervised baselines for all translation directions, improving the best performing unsupervised baseline with an average gain of 1.2 BLEU. In addition, it even outperforms all supervised baselines and achieves new state-of-the-art performance on Ne$\sri$En and En$\sri$Ne translations. It is impressive given that the supervised baselines \citet{Flores} and \citet{mbart} are very strong. Both methods are trained on around 600k parallel corpus and more than 70M monolingual corpora with supervised translation and iterative back-translation. Our method is also computationally efficient and easy to implement. As \mname offers a ready-to-use \llr$\sri$En NMT model, we only run back-translation once for building the reverse model. However, for the baselines (ID 2-3, 5-7), they run iterative back-translation for no less than two rounds and involve cross-translation, denoising autoencoding, or adversarial loss. They are much more complex and computational costly compared with our method. 

\subsection{Zero-shot Cross-lingual Generation}
\begin{table*}[t!]
  \centering 
  \resizebox{0.65\textwidth}{!}{
  \begin{tabular}{l|l|r|rrrrrr} 
  \toprule
  Model  &    Metric                      & En    & Hi    & Zh    &  Cs   & Nl    & Tr     &  Avg.    \\ \midrule
  \multirow{3}{*}{mBART-ft}                & ROUGE-1  & 41.5  &  16.4 &  19.8 & 29.8  & 35.2  & 32.2   &  26.7     \\ 
                                        & ROUGE-2  & 18.9  &   4.1 &   5.7 & 10.3  & 13.8  & 12.8   &  9.3        \\
                                        & ROUGE-L  & 35.5  &  15.0 &  17.7 & 26.1  & 30.5  & 28.2   &  23.5      \\ \midrule
 \multirow{3}{*}{\tabincell{l}{Ours w/o NMT\\pretraining}}       & ROUGE-1     & 40.5  &  35.8 &  32.7  & 33.7  & 37.2  & 40.6  &  36.0 \\
                                       & ROUGE-2     & 19.0  &  16.0 &  13.4  & 13.9  & 16.6  & 20.5  &  16.1     \\
                                       & ROUGE-L     & 35.2  &  31.4 &  28.6  & 29.7  & 32.5  & 35.9  &  31.6    \\
                                       \midrule
  \multirow{3}{*}{Ours}      & ROUGE-1  & 43.7  &  40.6 &  37.2 & 37.9  & 41.3  & 45.6   &  40.5      \\ 
                                        & ROUGE-2  & 21.5  &  20.1 &  16.4 & 17.4  & 20.1  & 25.3   &  19.9      \\
                                        & ROUGE-L  & 37.9  &  35.9 &  32.6 & 33.6  & 36.3  & 40.7   &  35.8      \\

  \bottomrule
  \end{tabular}}
  \caption{ROUGE results for zero-shot cross-lingual abstractive summarization. For ROUGE score, higher value is better. The `Avg' is the average score of all zero-shot directions.}
  \label{tab:xsum-result}
 \end{table*}  
In zero-shot generation with the source-side transfer, the NLG model is directly tested on unseen source languages during supervised training. As cross-lingual labeled data are scarce, such zero-shot generation is useful in the cross-lingual generation where the languages of input and output text are different. In this experiment, we focus on utilizing \mname for zero-shot cross-lingual abstractive summarization (ZS-XSUM). We believe such a framework can be easily extended to other zero-shot cross-lingual generation tasks.

The ZS-XSUM task is challenging, as we require the model to summarize (from document to abstract), translate (from input language to output language) and transfer (from auxiliary input language to target input language) at the same time. \mname already has the ability to translate and transfer, thus it offers a set of initialization parameters that can ease the learning of the ZS-XSUM model. Specifically, we initialize the ZS-XSUM model with \mname (1st)\footnote{Preliminary experiments show that this setting leads to slightly better performance than initialization with \mnamec.} and then train on labeled data of abstractive summarization with the TransF method. The trained model is tested on the cross-lingual summarization in a zero-shot manner where the source language is unseen during training.

\paragraph{Experiment Settings}

To build a strong ZS-XSUM model, we collect 1.2 million public document-summary pairs to form the training dataset, where the document is in the languages among En/De/Es/Fr/It/Pt/Ru and the summary is in En. We evaluate the performance on the Wikilingua dataset with Hi/Zh/Cs/Nl/Tr as source languages and English as the target language. All the test languages are unseen during training and validation. The dataset details are in the appendix. We compare the proposed method with the mBART-ft method which directly fine-tunes all mBART parameters and our proposed method in building \mname which is denoted as `Ours w/o NMT pretraining'.  

\paragraph{Results} As shown in Table~\ref{tab:xsum-result}, both of our methods outperform mBART-ft on all zero-shot directions by an average gain of 8.1 and 12.3 ROUGE-L. This is impressive given that mBART is a widely used MulPLM for the cross-lingual generation. We also observe that initializing with \mname is much better than XLM-R with the same TranF training method, demonstrating that the NMT pretraining is beneficial for the ZS-XSUM task. To build a cross-lingual generation model without labeled data, previous works usually resort to the translate-and-train or translate-and-test approaches or their extensions~\citep{shen2018zero,duan-etal-2019-zero}. For these approaches, an NMT system is required to translate either at the training or testing time. However, translate-and-train can only develop models for a few pre-specified source languages, while the decoding speed of translate-and-test is slow, especially for summarization where the input text is long. Besides, both approaches rely heavily on the performance of the NMT system. \mname shows that it is possible to build a strong universal cross-lingual NLG model that can support 100 source languages. This is promising, especially for low-resource languages which the NMT system translates poorly. Our model can also serve as a start point which can be further improved by fine-tuning on genuine or synthesized (produced by an NMT system) cross-lingual corpus. We leave more in-depth exploration as future work.

\section{Related Work}

\subsection{Multilingual Neural Machine Translation}
Early works on multilingual NMT show its zero-shot translation capability, where the tested translation direction is unseen during supervised training~\cite{johnson2017google,ha2016toward}. To further improve the zero-shot performance, one direction is to learn language-agnostic encoder representations and make the most of cross-lingual transfer. Some approaches modify the encoder architecture to facilitate language-independent representations. \citet{lu2018neural} incorporate an explicit neural interlingua after the encoder. \citet{liu2020improving,Chen2021ZeroshotCT} remove the residual connection at an encoder layers to relax the positional correspondence. Some other works introduce auxiliary training objectives to encourage similarity between the representations of different languages~\cite{Arivazhagan2019TheMI,AlShedivat2019ConsistencyBA,Pham2019ImprovingZT,pan-etal-2021-contrastive}. For example, \citet{pan-etal-2021-contrastive} utilize contrastive loss to explicitly align representations of a bilingual sentence pair. Recently, multilingual pretraining has demonstrated to implicitly learn language-agnostic representation~\cite{mbart,xlmr,hu20xtreme}. Inspired by this, some studies initialize multilingual NMT with the MulPLM or introducing the training objectives of MulPLM to multilingual NMT~\cite{Gu2019ImprovedZN,ji2020cross,mbart,Chen2021ZeroshotCT,GarciaXBT2}. Our work follows the last line but improves over them by making the most of MulPLM with a simple yet effective fine-tuning method and large-scale multilingual parallel dataset.

\subsection{Zero-shot Translation with Multilingual Pretrained Language Model }
For NLG tasks like neural machine translation, most work leverage multilingual pretrained seq2seq language models such as mBART~\cite{mbart}, mT5~\cite{xue2020mt5} and ProphetNet-X~\cite{qi-etal-2021-prophetnet} for cross-lingual transfer. For example, \citet{mbart} fine-tune mBART with the parallel dataset of one language pair and test on unseen source languages. Considering the great success of the multilingual pretrained encoder (MulPE) such as XLM-R~\cite{xlmr} and mBART~\cite{mbert} in zero-shot cross-lingual transfer for NLU tasks~\cite{hu20xtreme}, their use for cross-lingual transfer in NLG tasks is still under-explored. \citet{wei21on} fine-tunes their proposed MulPE to conduct zero-shot translation but use the \texttt{[CLS]} representation  as the encoder output. 

Our work is most similar to SixT~\cite{Chen2021ZeroshotCT}, as indicated by the name itself. However, since SixT focuses on designing a novel fine-tuning method, it conducts experiments with one auxiliary language, which heavily limits the model's performance. In addition, SixT only works on NMT, while SixT+ can not only perform translation but also serve as a pretrained model for various zero-shot cross-lingual generation tasks, such as low-resource NMT and cross-lingual abstractive summarization. 

\section{Conclusion}

In this paper, we introduce \mnamec, a strong many-to-English NMT model that supports 100 source languages but is trained once with the parallel dataset from only six source languages. Our model makes the most of cross-lingual transfer by initializing with XLM-R and conducting multilingual fine-tuning on the large-scale dataset with a simple yet effective two-stage training method. Extensive experiments demonstrate that \mname outperforms all baselines on many-to-English translation. When serving as a pretrained model, adding \mname initialization achieves new state-of-the-art performance for unsupervised NMT of low-resource and significantly outperforms mBART and XLM-R on zero-shot cross-lingual summarization.

\section*{Acknowledgements}
This project was supported by National Natural Science Foundation of China (No. 62106138) and Shanghai Sailing Program (No. 21YF1412100). Wenping Wang and Jia Pan acknowledge the support from Centre for Transformative Garment Production. We thank the anonymous reviewers for their insightful feedbacks on this work.

\bibliography{anthology,custom}
\bibliographystyle{acl_natbib}

\appendix

\section{Dataset} \label{sec:app-data}

\subsection{Machine Translation Dataset}
The AUX6 dataset is from WMT translation task and CCAligned corpus\footnote{\url{http://www.statmt.org/cc-aligned/}}. The validation and test sets are from newstest, WAT21 translation task\footnote{\url{http://lotus.kuee.kyoto-u.ac.jp/WAT/indic-multilingual/indic_wat_2021.tar.gz}}, IWSLT17 testset\footnote{\url{https://wit3.fbk.eu/2017-01-d}},  Flores Testset\footnote{\url{https://github.com/facebookresearch/flores/tree/main/floresv1}} and Tatoeba test sets\footnote{\url{https://object.pouta.csc.fi/Tatoeba-Challenge/test-v2020-07-28.tar}}. We use the first 20M sentence pairs of the CCAligned corpus for Es-En and Ru-En language pairs as training data. The Europarl De-En dataset is only used in the experiment of Table~\ref{tab:pos-enc}. All texts are tokenized by the same XLM-R sentencepiece \citep{sentencepiece} model. The source sentence length is limited to 512, which is the maximum source sentence length supported by XLM-R. More details are shown in Table~\ref{tab:app-traval-dataset} and Table~\ref{tab:app-test-dataset}. 

\begin{table}[th]
  \centering 
  \resizebox{0.9\columnwidth}{!}{
    \begin{tabular}{llll}
    \toprule
    Type          &   Lang          & Source          & \# Sent   \\ \midrule
    Training set  &   De-En         & Europarl v7     & 1.9M        \\ 
    Training set  &   De-En         & WMT19           & 41M      \\
    Training set  &   Es-En         & CCAligned       & 20M      \\ 
    Training set  &   Fi-En         & CCAligned       & 9.2M      \\ 
    Training set  &   Hi-En         & CCAligned       & 7.4M      \\ 
    Training set  &   Ru-En         & CCAligned       & 20M      \\ 
    Training set  &   Zh-En         & WMT18           & 22.6M      \\ \midrule
    Valid set     &   De-En         & Newstest 16     & 2999      \\ 
    Valid set     &   Es-En         & Newstest 10     & 2489      \\
    Valid set     &   Fi-En         & Newstest 19     & 1996      \\
    Valid set     &   Hi-En         & Newsdev 14      & 520      \\
    Valid set     &   Ru-En         & Newstest 16     & 2998     \\ 
    Valid set     &   Zh-En         & Newstest 17     & 2001      \\
    \bottomrule
    \end{tabular}}
  \caption{Training and valid set for many-to-English translation. `\# Sent' is the number of parallel sentences in the dataset.} 
  \label{tab:app-traval-dataset}
\end{table} 

\begin{table}[ht]
  \centering 
  \resizebox{0.9\columnwidth}{!}{
    \begin{tabular}{ll|ll}
    \toprule
    Lang   & Source        & Lang    & Source          \\ \midrule
    Ar-En  & IWSLT 17      & Lv-En   & Newstest 17     \\
    De-En  & Newstest 14   & My-En   & WAT21           \\ 
    Es-En  & Newstest 13   & Ne-En   & Flores v1         \\ 
    Et-En  & Newstest 18   & Nl-En   & Tatoeba         \\ 
    Fi-En  & Newstest 16   & Pl-En   & Newstest 20     \\ 
    Gu-En  & Newstest 19   & Ps-En   & Newstest 20    \\  
    Hi-En  & Newstest 14   & Ro-En   & Newstest 16     \\
    It-En  & Tatoeba       & Ru-En   & Newstest 20     \\
    Ja-En  & Newstest 20   & Si-En   & Flores v1         \\ 
    Kk-En  & Newtest 19    & Tr-En   & Newstest 16     \\ 
    Km-En  & Newstest 20   & Zh-En   & Newstest 18     \\ 
    Ko-En  & Tatoeba  \\
    \bottomrule
    \end{tabular}}
  \caption{Test sets for many-to-English translation.}
  \label{tab:app-test-dataset}
\end{table} 

\subsection{Unsupervised NMT dataset}

The monolingual dataset of Ne and Si consists of 7 million sentences that are sampled from CC100 \cite{xlmr} and CCNet \cite{ccnet} datasets. We select the best model with an unsupervised criterion based on the BLEU score of a `round-trip' translation following~\citep{lample2018unsupervised} by using 3000 monolingual Ne/Si sentences sampled from CC100 and CCNet datasets. The testsets of Ne and Si are from Flores testset \cite{Flores} \footnote{\url{https://github.com/facebookresearch/flores/tree/main/floresv1}}.

\subsection{Abstractive Summarization Dataset}
The training data of abstractive summarization task is from CNN/DailyMail,\footnote{\url{https://github.com/abisee/cnn-dailymail}} XSum,\footnote{\url{https://github.com/EdinburghNLP/XSum/tree/master/XSum-Dataset}} Wikihow\footnote{\url{https://github.com/mahnazkoupaee/WikiHow-Dataset}} and WikiLingua\footnote{\url{https://github.com/esdurmus/Wikilingua}} dataset. In total, the training set contains 1189k document-summary pairs. The average context length after performing sentencepiece is 669 tokens. We randomly sample 2000 Fr-En pairs and 3000 pairs for each test language from the WikiLingua dataset as the validation and test sets. As the maximum length of input tokens for XLM-R is 512, we just keep the first 512 tokens of context input if it is longer than 512. The model is evaluated on many-to-English abstractive summarization, where we summarize documents of various languages to English abstracts. More details are shown in Table~\ref{tab:app-xsum-dataset}.

\begin{table}[ht]
  \centering 
  \resizebox{0.4\textwidth}{!}{
    \begin{tabular}{ll|ll}
    \toprule
    Dataset   &  Lang pair  &  Source        &  \# Sent   \\ \midrule
    Train     &  En-En      & CNN/DailyMail  &  280K    \\
    Train     &  En-En      & XSum           &  204k     \\
    Train     &  En-En      & WikiHow        &  180K     \\
    Train     &  En-En      & WikiLingua     &  136K     \\
    Train     &  De-En      & WikiLingua     &  53K     \\
    Train     &  Es-En      & WikiLingua     &  106K     \\
    Train     &  Fr-En      & WikiLingua     &  59K     \\
    Train     &  It-En      & WikiLingua     &  46K     \\
    Train     &  Pt-En      & WikiLingua     &  77K     \\
    Train     &  Ru-En      & WikiLingua     &  48K     \\ \midrule
    Valid     &  Fr-En      & WikiLingua     &  2K     \\ \midrule
    Test      &  En-En      & WikiLingua     &  3K     \\
    Test      &  Cs-En      & WikiLingua     &  3K     \\
    Test      &  Hi-En      & WikiLingua     &  3K     \\
    Test      &  Nl-En      & WikiLingua     &  3K     \\
    Test      &  Tr-En      & WikiLingua     &  2.9K   \\
    Test      &  Zh-En      & WikiLingua     &  3K     \\
    \bottomrule
    \end{tabular}}
  \caption{Dataset for many-to-English abstractive summarization task.}
  \label{tab:app-xsum-dataset}
\end{table}
\begin{table}[ht]
  \centering 
  \resizebox{1.0\columnwidth}{!}{
    \begin{tabular}{lll|lll}
    \toprule
    ISO   & Language  & Family     & ISO   & Language    & Family   \\ \midrule
    ar  &  Arabic    &  Arabic     &  ko   & Korean     & Koreanic     \\       
    cs  & Czech      & Slavic      &  lv   & Latvian    & Baltic       \\      
    de  & German     & Germanic    &  my   &  Burmese   &  Sino-Tibetan   \\      
    en  & English    & Germanic    &  ne   & Nepali     & Indo-Aryan   \\     
    es  & Spanish    & Romance     &  nl   & Dutch      & Germanic     \\ 
    et  & Estonian   & Uralic      &  pl   &  Polish   &  Slavic       \\       
    fi  & Finnish    & Uralic      &  ps   &  Pashto   &  Iranian       \\     
    fr  & French     & Romance     &  ro   & Romanian   & Romance      \\      
    gu  & Gujarati   & Indo-Aryan  &  ru   & Russian    & Slavic       \\          
    hi  & Hindi      & Indo-Aryan  &  si   & Sinhala    & Indo-Aryan   \\        
    it  & Italian    & Romance     &  tr   &  Turkish  &  Turkic       \\
    ja  & Japanese   & Japonic     &  vi   & Vietnamese & Vietic       \\      
    kk  & Kazakh     & Turkic      &  zh   & Chinese    & Chinese      \\
    km  & Khmer      & Khmer       \\ 
    \bottomrule
    \end{tabular}}
  \caption{Languages used in this paper.}
  \label{tab:app-isocode}
\end{table} 

\section{Language Code}

We refer to the language information in Table 1 of \citet{fan2020m2m}. The languages used in this paper are shown in Table~\ref{tab:app-isocode}.

\section{Model and Training Details}

Since the \mname embeddings are initialized with XLM-R, all texts are tokenized with the same sentencepiece \citep[][SPM]{sentencepiece} tokenizer as XLM-R. The tokenizer is learned on the full Common Crawl data that includes 250k sentencepiece tokens. We do not apply additional preprocessing, such as true-casing or normalizing punctuation/characters. Following XLM-R, we add the \texttt{[BOS]} and \texttt{[EOS]} tokens at the head and tail of the input sentence, respectively.

\mname is trained on 128 Nvidia V100 GPUs (32GB) with 100k and 10k steps for the first and second training stage. The batch size is 4096 for each GPU. We use the Adam optimizer~\cite{adam} with $\beta_1=0.9$ and $\beta_2=0.98$. At the first stage, the learning rate is 0.0005 and the warmup step is 4000, while at the second stage, we set the learning rate as 0.0001 and do not use warmup. The dropout probabilities are set to be 0.1. All experiments are done with the fairseq toolkit~\citep{ott2019fairseq}.

\section{Comparison on the Many-to-many Translation}

The many-to-many \mname model is trained with AUX6 dataset using supervision from 12 translation directions. The m2m-100 model is the official 1.2B model\footnote{\url{https://github.com/pytorch/fairseq/tree/main/examples/m2m_100}} from \citet{fan2020m2m}. The results are shown in Table~\ref{tab:app-m2m-task-results}. 

\begin{table*}[t]
  \centering 
  \resizebox{1.0\textwidth}{!}{
    \begin{tabular}{l|l|rrrrrrrrrrrrr}
    \toprule
    \diagbox{Tgt}{Src}  & Model            & Nl   &  Ro   &  Sr   &  Lv   &  Pl   &  Ne   &  Gu   &  Ja   &  Mr   &  Kk   &  Km   &  Tr  &  Avg    \\ \midrule 
    \multirow{4}{*}{$\sri$ En} & m2m-100       & 29.7 &  40.7  &  39.6  &  33.1  &  27.1  &  13.2  &  1.7  & 23  & 22.3  & 5.2 &  14.0 &  33.0 &  23.6   \\
      & Ours (m2En)                          & 29.6  &  39.1  &  37.9  &  31.3  &  26.1  &  35.3  &  33.5  &  21.3  &  29.9  &  27.2  &  21.4  &  33.1  &  \tub{30.5}   \\
      & Ours (m2m)                            & 29.0 &  37.9  &  37.0  &  30.6  &  25.3  &  34.5  &  32.1  &  20.2  &  29.4  &  27.0  &  21.7  &  32.3  &  29.8    \\ 
     & Ours (m2m w/o PDE)                       & 29  &  38.2  &  37.4  &  30.3  &  25.5  &  34.2  &  32.4  &  20.6  &  29.5  &  26.8  &  21.4  &  32.4  &  29.8      \\ 
\midrule  
    \multirow{3}{*}{$\sri$ De}  & m2m-100      & 21.8 &  28.1  &  27.7  &  14.8  &  20.9  &  8.3   &   1.0  &  16.5 &  14.0 & 4.9 &  9.2 & 23.1 & 15.9      \\ 
     & Ours (m2m)                             & 20.1 &  24.5  &  24.1  &  19.8  &  17.5  &  16.0  &  15.1  &  11.9  &  14.5  &  14.8  &  11.8  &  18.7  &  \tub{17.4}    \\ 
     & Ours (m2m w/o PDE)                            & 19.3  &  24.4  &  23.4  &  19.1  &  17.4  &  15.3  &  14.1  &  11.1  &  13.6  &  13.6  &  11.6  &  17  &  16.7  \\ \midrule 
    \multirow{3}{*}{$\sri$ Es} &   m2m-100     & 18.9 &  24.1  &  22.6  &  20.7  &  19.8  &   8.6  &  1.8   & 15.8 &13.0 & 6.3 & 10.4 & 19.8 & 15.2     \\
    &   Ours (m2m)                            & 16.5 &  21.7  &  19.3  &  16.3  &  16.6  &  14.4  &  12.5  &  12.2  &  13.1  &  14.1  &  10.4  &  16.2  &  \tub{15.3}    \\ 
    &   Ours (m2m w/o PDE)                       & 16.7  &  21.8  &  19.1  &  15.9  &  16.4  &  14  &  11.7  &  11.2  &  12.2  &  13.5  &  10.3  &  15.6  &  14.9    \\ \midrule
    \multirow{3}{*}{$\sri$ Fi}  & m2m-100      & 14.4 &  18.0  &  17.6  &  17.4  &  14.6  &   6.2  &  1.1   & 11.5 & 9.1 & 4.4 & 7.0 & 14.3 & \tub{11.3}     \\
     & Ours  m2m                            & 11.6 &  13.8  &  12.7  &  12.4  &  11.1  &  10.0  &   8.7  &   7.4  &   8.3  &   9.0  &   7.3  &  10.0  &  10.2    \\
     & Ours (m2m w/o PDE)                     & 11.5  &  13.2  &  12.2  &  12.3  &  10.8  &  9.5  &  7.9  &  6.5  &  7.8  &  8.5  &  6.9  &  9.6  &  9.7     \\ \midrule
    \multirow{3}{*}{$\sri$ Hi}   &  m2m-100    & 16.1 &  20.7  &  20.6  &  18.8  &  16.1  &  11.1  &  1.4   & 14.8 & 18.2  & 3.7  & 8.0  & 19.1 & 14.1  \\
      & Ours (m2m)                             & 14.5 &  18.2  &  18.3  &  15.3  &  13.4  &  20.0  &  20.0  &  10.8  &  17.7  &  13.3  &  10.7  &  14.2  &  15.5\\ 
    & Ours (m2m w/o PDE)                    & 14.9  &  18.4  &  18.3  &  15.2  &  13.6  &  21.1  &  20.4  &  11.1  &  18.3  &  13.8  &  9.8  &  14.5  &  \tub{15.8}      \\ \midrule
    \multirow{3}{*}{$\sri$ Ru} & m2m-100       & 17.2 &  24.4  &  25.0  &  19.1  &  18.6  &  7.4   &   1.0  &  14.4  &12.5  &4.8 &8.5  &18.5  &14.3    \\
     & Ours  (m2m)                            & 13.9 &  19.4  &  20.6  &  19.4  &  16.0  &  13.0  &  12.7  &   9.9  &  12.0  &  14.5  &   9.8  &  14.5  &  \tub{14.6} \\ 
 & Ours (m2m w/o PDE)                 & 14.2  &  19.2  &  20  &  18.9  &  15.6  &  12.6  &  11.9  &  9.2  &  11.1  &  13.7  &  9.4  &  13.7  &  14.1     \\ \midrule
   \multirow{3}{*}{$\sri$ Zh} & m2m-100        & 25.7 & 29.4   &  29.2  &  22.6  &  25.5  & 12.8   & 0.7    & 26.9& 19.5 & 7.3 & 12.4 & 26.7 & 19.9     \\
    & Ours (m2m)  & 26.3                       & 29.6 &  28.7  &  27.1  &  25.2  &  24.6  &  22.5  &  24.7  &  23.1  &  23.9  &  20.3  &  26.0  &  \tub{25.2}  \\
    & Ours (m2m w/o PDE)  & 26.3  &  29.4  &  28.6  &  26.9  &  25  &  23.8  &  21.9  &  24.5  &  22.6  &  23.5  &  19.5  &  25.9  &  24.8  \\
    \bottomrule
    \end{tabular}}
  \caption{BLEU comparison of our many-to-many NMT model (\mname m2m) with m2m-100 on zero-shot translations. We use a target-language-aware linear projection layer to generate different target languages for the \mname m2m model. Ours (m2En) is the many-to-English \mname model trained with the AUX6 dataset. We include the result of \mname m2m w/o PDE to help study the effect of PDE. The best average BLEU for each target language is bold and underlined. }
  \label{tab:app-m2m-task-results}
\end{table*}

\section{Effect of Positional Disentangled Encoder}

We compare the \mname with and without (w/o) positional disentangled encoder (PDE) on different training datasets: Europarl (1.9M), WMT19 (41M), and AUX6 (120M). The results are shown in Table~\ref{tab:app-pos-enc}. We also conduct experiments on \mname m2m, as shown in Table~\ref{tab:app-m2m-task-results}.

\begin{table*}[t]
  \centering 
  \resizebox{0.9\textwidth}{!}{
  \begin{tabular}{lc|r|rrrrrrrrrrr}
  \toprule
  \#Sent & Config.               & De    &  Nl  &  Ro  &  It  &  Lv  &  Et  &  Ne  &  Si  &  Gu  &  Ja  &  Ko  &  Avg.      \\ \midrule
   \multirow{2}{*}{1.9M} & Ours     & 28.7  &  44.7  &  28.3  &  39.2  &  16.0  &  21.4  &  11.0  &  10.0  &  12.8  &   8.0   &  23.5  & \tub{21.5}     \\ 
    & w/o PDE                         & 29.1  &  44.2  &  27.2  &  39.0  &  15.3  &  20.5  &  10.1  &   8.8  &  12.6  &   7.1   &  20.1  & 20.5  \\  \midrule
   \multirow{2}{*}{41M} & Ours      & 33.8  &  54.7  &  33.9  &  43.0  &  19.7  &  25.7  &  14.4  &  12.2  &  17.3  &  10.7   &  31.2  & \tub{26.3}    \\ 
   &  w/o PDE                         & 34.1  &  54.9  &  33.5  &  43.5  &  19.7  &  25.5  &  14.1  &  12.0  &  17.0  &  10.3   &  30.2  & 26.1  \\ \midrule
    \multirow{2}{*}{120M}  & Ours   & 35.3  &  58.5  &  38.6  &  60.9  &  23.3  &  30.5  &  23.7  &  17.5  &  27.5  &  13.1   &  33.3  &  \tub{32.9}       \\ 
   &  w/o PDE                         & 35.2  &  58.5  &  39.0  &  61.1  &  23.2  &  30.1  &  23.6  &  17.4  &  27.2  &  13.7   &  32.5  &  \tub{32.9}    \\
  \bottomrule
  \end{tabular}}
  \caption{The BLEU comparison between \mname with and without positional disentangled encoder (PDE). The best average BLEU for each training dataset is bold and underlined.}
  \label{tab:app-pos-enc}
\end{table*}

\section{Unsupervised NMT with \mnamec}
In addition to CRISS and m2m-100, we compare with the state-of-the-art unsupervised and supervised baselines from the literature on these two languages. Most of these additional baselines are not multilingual and are explicitly designed for low-resource language translation. 

\noindent $\small \bullet$ Unsupervised baselines. We include the results of three unsupervised methods. \citet{Flores} utilize Hi as auxiliary language and train with auxiliary supervised translation and iterative back-translation. \citet{GarciaXBT2} utilize six languages as auxiliary languages and present a three-stage method with various loss functions, including auxiliary supervised translation, iterative back-translation, denoising autoencoding and cross translation. \citet{ko2021adapting} fine-tune mBART on the parallel dataset from Hi and monolingual data in an iterative manner with auxiliary supervised translation, back-translation, denoising autoencoding and adversarial objective. Note that these methods utilize much more monolingual data than ours.

\noindent $\small \bullet$ Supervised baselines. We report the supervised results in mBART~\cite{mbart} and the FLoRes dataset benchmarks~\cite{Flores} for reference. These two methods are very strong. Both methods are trained on around 600k parallel corpus and more than 70M monolingual corpora with supervised translation and iterative back-translation. \citet{mbart} initialize the model with mBART while \citet{Flores} use auxiliary parallel corpus from related language for the $\texttt{Ne}{\leftrightarrow}\texttt{En}$ translations. 

\section{XTREME benchmark results}

All models are evaluated on the XTREME benchmark \cite{hu20xtreme} with jiant toolkit\footnote{\url{https://github.com/nyu-mll/jiant}}. We follow the same settings with \citet{phang2020english} for fine-tuning and testing. The detailed results for each languages on each task are shown in Table~\ref{tab:full_xnli} to Table~\ref{tab:full_tatoeba}.

\begin{table*}[ht]
  \resizebox{\textwidth}{!}{\small
  \begin{tabular}{lcccccccccccccccc}
  \toprule
        & ar & bg & de & el & en & es & fr & hi & ru & sw & th & tr & ur & vi & zh & Avg \\  \midrule
  XLM-R & 77.2  &  83  &  82.5  &  80.8  &  88.7  &  83.7  &  82.2  &  75.6  &  79.1  &  71.2  &  77.4  &  78  &  71.7  &  79.3  &  78.2  &  79.2 \\
  \ftall  &  72.3  &  81.3  &  81.6  &  76.3  &  86.7  &  81.9  &  80.3  &  74.0  &  78.5  &  58.0  &  72.7  &  73.1  &  67.0  &  77.4  &  77.9  &  75.9  \\ 
  \mEnres &  77.2  &  81.9  &  82.3  &  80.1  &  87.5  &  83.0  &  82.0  &  75.1  &  78.5  &  69.8  &  75.0  &  77.8  &  70.2  &  78.6  &  78.4  &  78.5  \\
  \mmres  &  79.1  &  83.3  &  83.4  &  82.3  &  88.6  &  84.2  &  83.4  &  76.9  &  80.2  &  71.3  &  77.2  &  78.5  &  72.1  &  80.0  &  79.5  &  80.0    \\
  \bottomrule  
  \end{tabular}
  }
  \caption{Full XNLI Results (accuracy)}
  \label{tab:full_xnli}
  \end{table*}

  \begin{table*}[ht]
  \centering\small
  \begin{tabular}{lcccccccc}
  \toprule
  & de & en & es & fr & ja & ko & zh & Avg \\  \midrule
  XLM-R   & 89.7  &  94.7  &  90.1  &  90.4  &  78.7  &  79.0  &  82.3  &  86.4  \\
  \ftall   &  89.1  &  95.3  &  90.0  &  89.9  &  77.9  &  76.6  &  82.8  &  85.9 \\
  \mEnres  &  91.0  &  95.9  &  90.9  &  91.2  &  81.2  &  81.5  &  84.6  &  88.0  \\ 
  \mmres   &  90.8  &  95.0  &  91.4  &  91.2  &  82.8  &  81.8  &  84.8  &  88.3  \\
  \bottomrule
  \end{tabular}
  \caption{Full PAWS-X Results (F1 score)}
  \label{tab:full_pawsx}
  \end{table*}

  \begin{table*}[t!]
    \resizebox{\textwidth}{!}{\small
    \begin{tabular}{lccccccccccccccccc}
    \toprule
    & af & ar & bg & de & el & en & es & et & eu & fa & fi & fr & he & hi & hu                                                                         \\  \midrule
    XLM-R  &  89.8  &  67.5  &  88.1  &  88.5  &  86.3  &  96.1  &  88.3  &  86.5  &  72.5  &  70.6  &  85.8  &  45.1  &  68.3  &  76.4  &  82.6   \\
    \ftall   & 83.8  &  57.8  &  80.8  &  79.0  &  75.3  &  95.9  &  72.0  &  78.9  &  57.6  &  60.2  &  74.8  &  75.3  &  61.7  &  58.9  &  74.8  \\
    \mEnres  & 89.8  &  69.4  &  89.5  &  89.4  &  86.9  &  96.0  &  87.5  &  86.9  &  72.7  &  70.1  &  86.9  &  86.5  &  71.9  &  70.5  &  83.8  \\
    \mmres   & 87.1  &  64.7  &  87.6  &  86.3  &  86.0  &  95.2  &  86.9  &  85.8  &  72.6  &  66.5  &  84.8  &  84.2  &  69.0  &  75.0  &  81.0  \\  \midrule  \midrule
             & id & it         & ja  & ko & mr & nl & pt & ru & ta & te &tr & ur & vi  & zh & Avg \\  \midrule
     XLM-R   & 72.4  &  89.4  &  15.9  &  53.9  &  80.8  &  89.5  &  87.6  &  89.5  &  65.2  &  86.6  &  76.3  &  70.3  &  56.8  &  25.7  &  74.2  \\
    \ftall   &  68.6  &  72.6  &  17.6  &  42.6  &  71.4  &  85.6  &  78.2  &  76.8  &  60.1  &  77.6  &  68.5  &  56.6  &  49.8  &  34.0  &  67.1 \\ 
    \mEnres  &  72.3  &  87.3  &  33.5  &  52.6  &  81.0  &  89.6  &  85.9  &  89.8  &  64.4  &  84.8  &  76.5  &  61.6  &  56.1  &  34.5  &  76.1 \\ 
    \mmres   &  72.4  &  86.6  &  19.3  &  50.9  &  82.4  &  88.3  &  85.8  &  87.7  &  61.7  &  87.7  &  76.0  &  69.2  &  57.0  &  19.7  &  74.4  \\   \bottomrule
    \end{tabular}
    }
    \caption{Full POS Results (F1 score)}
    \label{tab:full_udpos}
\end{table*}
  
  \begin{table*}[t!]
  \resizebox{\textwidth}{!}{\small
  \begin{tabular}{lccccccccccccccccccccc}
  \toprule
  & af & ar & bg & bn & de & el & en & es & et & eu & fa & fi & fr & he & hi & hu & id & it & ja & jv & ka \\  \midrule
  XLM-R  & 78.9  &  53.0  &  81.4  &  78.8  &  78.8  &  79.5  &  84.7  &  79.6  &  79.1  &  60.9  &  61.9  &  79.2  &  80.5  &  56.8  &  73.0  &  79.8  &  53.0  &  81.3  &  23.2  &  62.5  &  71.6  \\
  \ftall   & 71.6  &  37.6  &  65.9  &  53.8  &  61.9  &  44.5  &  82.7  &  67.5  &  64.8  &  44.1  &  32.5  &  65.1  &  76.4  &  39.4  &  58.3  &  67.9  &  52.4  &  75.4  &  13.4  &  53.2  &  52.7 \\
  \mEnres  & 74.4  &  52.2  &  76.7  &  70.1  &  76.4  &  75.8  &  82.6  &  74.0  &  74.3  &  61.9  &  50.7  &  76.1  &  78.4  &  52.4  &  67.2  &  76.8  &  55.5  &  79.6  &  19.7  &  61.7  &  62.4  \\
  \mmres   & 75.5  &  44.7  &  77.1  &  67.4  &  78.4  &  72.2  &  80.2  &  68.7  &  75.2  &  62.0  &  50.1  &  78.9  &  77.2  &  46.8  &  66.9  &  76.6  &  50.3  &  76.7  &   9.8  &  58.9  &  57.3  \\  
  \midrule
  \midrule
  & kk & ko & ml & mr & ms & my & nl & pt & ru & sw & ta & te & th & tl & tr & ur & vi & yo & zh & Avg & \\  \midrule
  XLM-R   &  56.2  &  60.0  &  67.8  &  68.1  &  57.1  &  54.3  &  84.0  &  81.9  &  69.1  &  70.5  &  59.5  &  55.8  &  1.3  &  73.2  &  76.1  &  56.4  &  79.4  &  33.6  &  33.1  &  65.4  \\
  \ftall   &  33.4  &  23.0  &  41.5  &  44.8  &  68.5  &  40.0  &  77.7  &  76.3  &  52.6  &  63.0  &  40.1  &  34.9  &  2.2  &  73.1  &  71.4  &  42.2  &  65.2  &  32.5  &  19.1  &  52.1 \\
  \mEnres  &  52.7  &  54.5  &  54.8  &  56.0  &  69.8  &  45.3  &  80.8  &  80.4  &  67.1  &  62.6  &  52.3  &  46.8  &  0.5  &  72.2  &  77.7  &  66.7  &  74.1  &  45.3  &  27.8  &  62.2  \\
  \mmres   &  50.0  &  49.1  &  52.6  &  55.5  &  73.1  &  47.3  &  81.2  &  78.7  &  52.4  &  59.1  &  50.2  &  44.0  &  1.4  &  71.3  &  75.7  &  48.4  &  73.7  &  33.5  &  10.1  &  59.0  \\
  \bottomrule
  \end{tabular}
  }
  \caption{Full NER Results (F1 score)}
  \label{tab:full_panx}
  \end{table*}
  
  \begin{table*}[t!]
  \resizebox{\textwidth}{!}{\small
  \begin{tabular}{llcccccccc}
  \toprule
  & ar & de & en & es & hi & vi & zh & Avg \\  \midrule
  XLM-R  & 66.6 / 47.1  &  70.1 / 54.9  &  83.5 / 70.6  &  74.1 / 56.6  &  70.6 / 53.1  &  74 / 52.9  &  62.1 / 37.0  &  71.6 / 53.2 \\ 
  \ftall   & 54.8 / 35.3  &  63.6 / 47.2  &  80.1 / 66.8  &  68.6 / 48.9  &  51.7 / 31.3  &  66.2 / 45.2  &  55.1 / 33.6  &  62.9 / 44.0  \\ 
  \mEnres  & 62.6 / 40.8  &  67.9 / 51.0  &  80.2 / 65.7  &  71.4 / 52.5  &  66.1 / 46.7  &  71.1 / 49.1  &  61.8 / 36.4  &  68.7 / 48.9  \\ 
  \mmres   & 65.2 / 44.6  &  70.5 / 55.3  &  82.1 / 68.4  &  74.1 / 55.6  &  69.5 / 50.5  &  73.0 / 51.1  &  60.7 / 36.2  &  70.7 / 51.7  \\ 
  \bottomrule
  \end{tabular}
  }
  \caption{Full MLQA Results (F1 / EM score)}
  \label{tab:full_mlqa}
  \end{table*}
  
\begin{table*}[t!]
  \centering\small
  \begin{tabular}{lccccc}  \toprule
  & de & fr & ru & zh & Avg \\  \midrule
  XLM-R  & 66.5  &  73.5  &  56.7  &  67.5  &  66.0  \\
  \ftall   &  3.4  &   8.0  &   2.4  &  17.9  &   7.9 \\
  \mEnres  & 89.6  &  84.1  &  86.6  &  83.1  &  85.9 \\ 
  \mmres   & 91.8  &  86.5  &  88.4  &  85.4  &  88.0 \\ 
  \bottomrule
  \end{tabular}
  \caption{Full BUCC Results (F1 Score)}
  \label{tab:full_bucc2018}
  \end{table*}
  
\begin{table*}[t!]
  \resizebox{\textwidth}{!}{\small
  \begin{tabular}{lccccccccccccccccccc}  \toprule
  & af & ar & bg & bn & de & el & es & et & eu & fa & fi & fr & he & hi & hu & id & it & ja & jv \\  \midrule
  XLM-R  & 58.2  &  47.5  &  71.6  &  43.0  &  88.8  &  61.8  &  75.7  &  52.2  &  35.8  &  70.5  &  71.6  &  73.7  &  66.4  &  72.2  &  65.4  &  77.0  &  68.3  &  60.6  &  14.1  \\
  \ftall   &  22.1  &  56.6  &  80.5  &  55.8  &  96.2  &  14.6  &  93.0  &  60.2  &  14.8  &  72.1  &  92.0  &  65.7  &  68.2  &  92.0  &  49.7  &  52.3  &  50.2  &  64.3  &  5.4 \\  
  \mEnres  &  74.4  &  72.8  &  87.2  &  74.6  &  98.1  &  83.1  &  96.1  &  81.5  &  54.6  &  91.0  &  94.6  &  90.7  &  82.0  &  94.2  &  86.9  &  91.9  &  87.9  &  91.1  & 19.5 \\
  \mmres   &  65.6  &  76.4  &  88.8  &  74.8  &  98.1  &  83.8  &  96.8  &  80.4  &  54.1  &  92.5  &  94.9  &  87.1  &  84.5  &  94.4  &  87.1  &  92.1  &  84.4  &  92.2  & 16.1  \\
  \midrule
  \midrule
  & ka & kk & ko & ml & mr & nl & pt & ru & sw & ta & te & th & tl & tr & ur & vi & zh & Avg & \\  \midrule
  XLM-R  & 52.1  &  48.5  &  61.4  &  65.4  &  56.8  &  80.8  &  82.2  &  74.1  &  20.3  &  26.4  &  35.9  &  29.4  &  36.7  &  65.7  &  24.3  &  74.7  &  68.3  &  57.7 \\
  \ftall   & 62.1  &  44.0  &  39.2  &  76.7  &  71.3  &  55.6  &  78.2  &  89.2  &  17.2  &  59.3  &  68.8  &  81.2  &  11.7  &  55.9  &  66.0  &  69.5  &  91.0  &  59.5 \\ 
  \mEnres  & 77.6  &  68.0  &  86.4  &  91.9  &  83.6  &  93.3  &  93.7  &  90.8  &  22.8  &  74.9  &  85.9  &  91.4  &  55.8  &  90.4  &  84.0  &  93.9  &  94.2  &  81.4\\
  \mmres   & 83.0  &  69.6  &  88.9  &  93.6  &  84.4  &  91.4  &  93.4  &  91.5  &  20.0  &  80.1  &  87.6  &  91.4  &  52.7  &  87.1  &  83.5  &  94.8  &  94.7  &  81.4 \\
  \bottomrule
  \end{tabular}
  }
  \caption{Full Tatoeba Results (Accuracy)}
  \label{tab:full_tatoeba}
\end{table*}

\end{document}